\pdfoutput=1
\documentclass[11pt]{article}

\usepackage[preprint]{acl}
\usepackage{times}
\usepackage{latexsym}
\usepackage{booktabs}
\usepackage{tabularx}
\usepackage{multirow}
\usepackage{enumitem}
\usepackage{adjustbox}
\usepackage{microtype}
\usepackage{inconsolata}
\usepackage{graphicx}
\usepackage{amsmath}
\usepackage[T2A,T1]{fontenc}
\usepackage[utf8]{inputenc}
\usepackage[ukrainian, main=english]{babel}

\title{Graph-Based Detection of Disinformation Narrative Diffusion between Russian and Ukrainian Telegram Channels}

\author{
  Yuliia Vistak \\
  Ukrainian Catholic University \\
  \texttt{vistak.pn@ucu.edu.ua}
  \And
  Viktoriia Makovska \\
  Ukrainian Catholic University \\
  \texttt{makovska.pn@ucu.edu.ua}
  \AND
  Vera Schmitt \\
  Technische Universität Berlin \\
  \texttt{vera.schmitt@tu-berlin.de}
  \And
  Veronika Solopova \\
  Technische Universität Berlin \\
  \texttt{veronika.solopova@tu-berlin.de}
}

\begin{document}
\selectlanguage{english}
\maketitle
\begin{abstract}
Detecting disinformation narratives on social media is challenging due to the scale of amplification, rapid evolution, and linguistic variability of online content. We propose a graph-based framework for identifying and analyzing disinformation narratives in Telegram ecosystems by combining weak supervision with propagation graph analysis. The approach aggregates semantically related claims into narrative-level clusters and models their diffusion across interconnected channels. This enables the detection of coordinated narrative amplification that is difficult to capture through post-level analysis alone. Our results demonstrate that integrating textual signals with network structure provides a scalable method for detecting disinformation narratives and offers insights into how they propagate within large-scale messaging environments.
\end{abstract}

\section{Introduction}

Disinformation at scale remains a persistent challenge for modern information ecosystems, with content volumes far exceeding the capacity of manual verification and fact-checking \citep{wardle2017information}. This challenge is especially acute in conflict settings, where disinformation evolves rapidly and is repackaged across platforms and communities. That's why narrative-level representation offers a practical abstraction for analyzing high-volume environments: rather than tracking individual posts and factuality of their claims, they aggregate semantically related claims into higher-level frames that remain stable despite linguistic variation and cross-platform adaptation \cite{nikolaidis-etal-2025-polynarrative}. Prior work in computational propaganda and fake news analysis shows that rhetorical framing and narrative structure capture systematic signals that generalize beyond surface text, enabling scalable trend detection across large corpora \citep{rashkin2017truth}.

\begin{figure}
    \centering
    \includegraphics[width=1\linewidth]{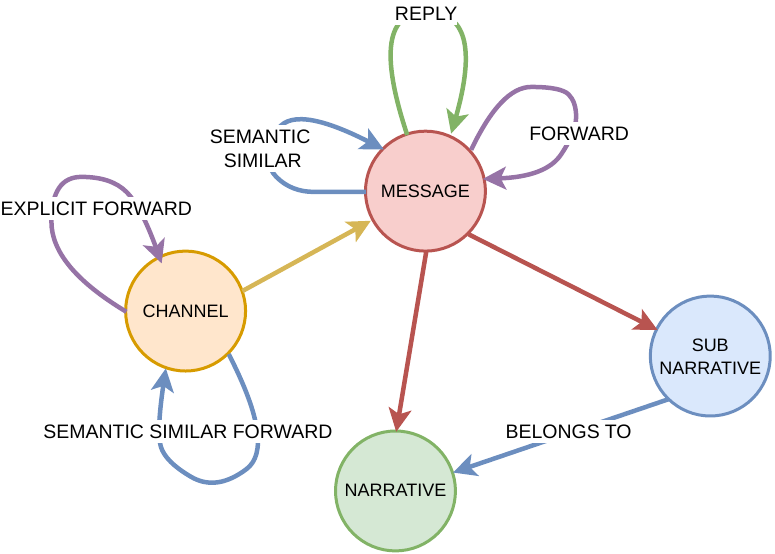}
    \caption{Schema of the graph used in our analysis, with channel, message, narrative, and sub-narrative nodes, and edges for posting, reply, explicit forwarding, semantic similarity, and narrative assignment.}
    \label{fig:graph_schema}
\end{figure}

Telegram is a particularly important platform for studying these dynamics, with a particularly high share of media space in Eastern Europe \cite{makhortykh2025evolutionwartimediscoursetelegram}. With moderation enforcement increased on mainstream social media in the early 2020s, harmful and conspiratorial communities migrated toward lower-moderation spaces such as Telegram \citep{kalkbrenner2025misinfotelegraph}. Its channel-based broadcast architecture and native forwarding mechanism make information diffusion \emph{structurally observable}: forwarded/forwarded-from metadata enables reconstruction of cross-channel propagation networks \citep{lamorgia2025tgdataset}. Telegram is also a central venue for Kremlin-related and anti-Kremlin communications during the Russia-Ukraine conflict, operating at substantial scale \citep{bawa2025telegram}. Despite this, misinformation detection research remains heavily skewed toward English-language and platforms such as X \citep{kalkbrenner2025misinfotelegraph}. Content-only classifiers are often brittle across languages and domains, while disinformation is frequently better characterized by amplifiers (e.g., botnets, troll farms, and coordinated sockpuppet accounts) and their patterns, rather than by the lexicon of the messages. Propagation-aware methods exploit the empirical observation that deceptive and credible information spreads differently, thereby providing more language-agnostic and manipulation-resistant signals \citep{monti2019fake}.
Therefore, in this study, we pursue network-driven disinformation narrative analysis for the Ukrainian Telegram ecosystem, building on MisinfoTeleGraph \citep{kalkbrenner2025misinfotelegraph} while adapting it to a conflict-specific, narrative-centric setting. We collect posts from a manually curated set of Telegram channels and construct a directed \emph{share graph} based on cross-channel forwarding (\autoref{fig:graph_schema}). Our supervision target is \emph{multi-class narrative assignment}: each message may activate multiple disinformation narratives. We operationalize the label space extending narrative taxonomy from the VoxCheck Propaganda Diary (database of narratives, created by fact-checking organization VoxCheck) and match messages against a fine-grained inventory of \textbf{380} pro-Russian disinformation subnarratives.~\footnote{\url{https://russiandisinfo.voxukraine.org/}} \footnote{\url{https://voxukraine.org/en/voxcheck}}

Methodologically, we compare three families of weak labeling approaches: semantic similarity to narrative descriptions, multilingual NLI-based zero-shot classification, and instruction-tuned LLM zero-, few-shot prompting. These methods are evaluated on a seed set of human-labeled messages, with model selection or ensembling guided by precision--coverage trade-offs. The resulting weak labels are attached to message nodes and, where needed, aggregated to the channel level. This labeled content is then analyzed in two complementary graph layers: an explicit share graph induced by Telegram forwarding metadata, and a semantic propagation graph induced by cross-channel message similarity. The full codebase, including labeling pipelines, graph construction, and evaluation scripts, is publicly available at GitHub repository.~\footnote{\url{https://github.com/yuliavistak/TeleNarratives.git}}
\section{Background and Related Work}
\label{sec:related-work}

This section reviews the main strands of work that inform our approach: disinformation narrative analysis, weak supervision for large-scale labeling, and graph-based modeling of information spread. Together, these lines of research motivate our focus on narrative-level detection in Telegram and our combination of weak labeling with graph-based propagation analysis.

\subsection{From Claims to Narratives}

Narratives are recurring ideological structures that organize multiple claims into coherent worldviews \citep{Hameleers2023}. Their effectiveness lies not in factual accuracy but in emotional resonance and identity alignment, which makes them resilient to corrective information \citep{Dahlstrom2014}. As a result, disinformation detection increasingly focuses on identifying narrative patterns rather than isolated falsehoods.
Prior work explicitly models narratives as hierarchical or multi-level structures. PolyNarrative introduces a multilingual dataset with coarse- and fine-grained narrative labels \citep{nikolaidis-etal-2025-polynarrative}, while datasets such as EUvsDisinfo provide benchmarks for detecting pro-Kremlin EU-targeting disinformation narratives in news articles \citep{10.1145/3627673.3679167}. Domain-specific studies have also examined narrative ecosystems in contexts such as climate change denial \citep{rojas2024acards, upravitelev2026retrievingclimatechangedisinformation}. We build on this by adopting a hierarchical narrative representation tailored to the Ukrainian information environment.

\subsection{Weak Supervision for Narrative Detection}

Expert annotation remains a bottleneck for narrative-level disinformation detection. Programmatic Weak Supervision addresses this challenge by combining multiple noisy labeling sources (known as labeling functions) to generate scalable, probabilistic supervision \cite{ratner2017snorkel}. Such sources may include heuristic rules, semantic similarity, or outputs of zero-shot classifiers.

In multilingual settings, weak labeling functions commonly rely on semantic similarity between texts and narrative descriptions, natural language inference (NLI)–based entailment scoring, or zero-shot classification with instruction-tuned language models \citep{yin-etal-2019-benchmarking, schick-schutze-2021-exploiting, ratner2017snorkel}. Weak supervision is well-suited for Telegram, where narrative inventories can be externalized and calibrated using small expert-labeled seed sets \citep{kalkbrenner2025misinfotelegraph}.

\subsection{Disinfo Spreading Networks Analysis via Graphs}

Graph-based approaches model misinformation not only through message content, but also through the structure of its spread. Early work showed that credibility can be inferred in part from diffusion and interaction patterns in social media streams \citep{castillo2011information}. Large-scale evidence from Twitter demonstrated that false news spreads faster, farther, and more broadly than true news, which motivated propagation-aware approaches as an alternative to content-only classification \citep{vosoughi2018spread}. Building on this, later work explicitly represented misinformation spread as graphs \citep{monti2019fake, bian2020rumor}.
Our work shifts the emphasis from message- or story-level misinformation detection to \emph{narrative-level} propagation analysis in Telegram.
\section{Methodology}
\label{sec:methodology}

Our methodology consists of four main components: (i) data collection, (ii) definition of a multi-label narrative assignment task using VoxCheck subnarratives, (iii) weak labeling via multiple few-shot and similarity-based labeling functions, and (iv) share graph construction from Telegram forwarding metadata.

\subsection{Data Collection}
\label{sec:data_collection}
We manually compiled a set of public Telegram channels relevant to war-related political discourse. The channel selection heavily relied on data from the fact-checking organization such as Spravdi.~\footnote{\href{https://spravdi.org/spilna-zayava-z-pryvodu-zahystu-informaczijnogo-prostoru-ukrayiny-vid-rosijskyh-vorozhyh-telegram-kanaliv/}{Spravdi channel list}.} We began with channels they identified as “unreliable” or “suspicious” and supplemented these with popular Ukrainian news channels exceeding 300,000 followers. This resulted in a total of \textbf{98} channels. Although we recognize that any curated list involves some bias, our aim was to make the final dataset as balanced as possible.

We also adopt the \textbf{VoxCheck Propaganda Diary}, which organizes pro-Russian disinformation into a structured taxonomy of \textbf{26} high-level \textbf{narratives} and \textbf{360} fine-grained \textbf{sub-narratives}, identified and curated during 2022--2023. Some of these narratives directly relate to the Russian--Ukrainian conflict (e.g. ``The war in Ukraine demonstrates the supremacy of Russian weapons''), while others address broader societal allegations (e.g. ``The Russian language is being suppressed in Ukraine'').
This taxonomy reflects real-world monitoring practices and enables multi-class narrative assignment, capturing the fact that individual messages may activate multiple narrative elements simultaneously.   

The messages dataset includes (i) message text, (ii) timestamps, (iii) channel and message identifiers, and (iv) native forwarding provenance fields (\texttt{is\_forwarded}, \texttt{fwd\_from\_channel\_id}, \texttt{fwd\_from\_message\_id}). This design tracks how content spreads across channels without using any private user data.

\begin{figure}[ht]
    \centering
    \includegraphics[width=0.5\textwidth]{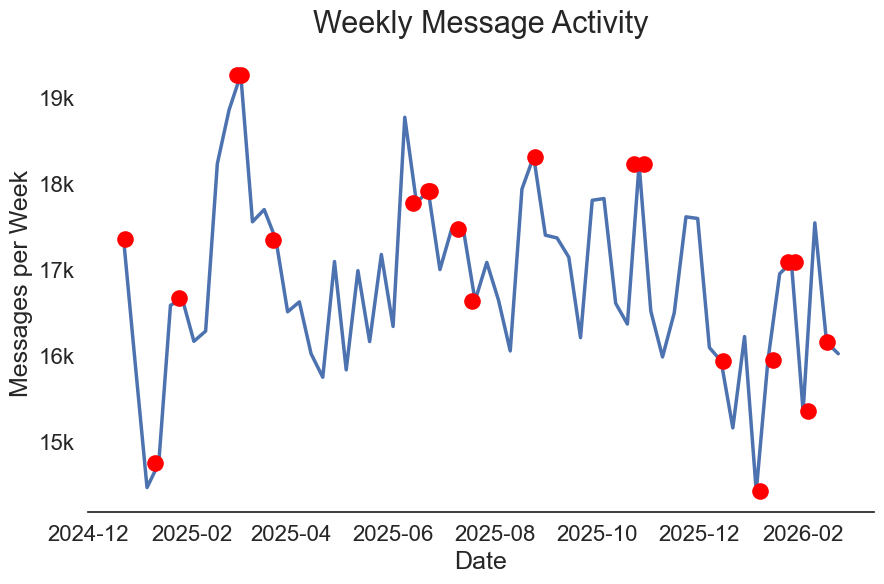}
    \caption{Weekly message activity within the curated Telegram corpus. Red markers denote significant political or conflict-related events (see Table ~\ref{tab:events} in the appendix for a comprehensive list). The peaks in message volume frequently coincide with these events.}
    \label{fig:weekly_temporal_analysis}
\end{figure}

The resulting messages corpus is primarily bilingual, containing messages in both \textbf{Ukrainian} and \textbf{Russian}.
Our primary objective was to analyze the discourse throughout 2025 and to include the early months of 2026, up to the date of our final data extraction. Consequently, the dataset consists of \textbf{1,352,668 messages} published between \textbf{16 December 2024} and \textbf{27 February 2026}. Figure \ref{fig:weekly_temporal_analysis} shows the temporal distribution of messages during the observation period.

To investigate the propagation of specific narratives and the underlying network topology between channels, we focused our analysis on the subset of messages involved in forwarding chains. This includes both \textbf{forwarded messages} and the \textbf{original posts} from which content was forwarded. From the initial corpus, this refined subset consists of \textbf{70,595} messages.

Some channels do not use the official ``forward'' button, copying and pasting text with only tiny changes. We call this \textit{implicit forwarding.} To catch these cases, we looked for messages with very similar meanings. For that reason, the decision was to transform messages into embedding vectors via \texttt{baai/bge-m3}, compute the similarity score between them, and take pairs with high score in further analysis. \footnote{\url{https://huggingface.co/BAAI/bge-m3}} 

This step enabled the inclusion of \textbf{10,872} semantically similar pairs (representing \textbf{10,774} additional messages). By tracking both official forwards and these ``copy-paste'' reposts, we get a much clearer picture of how news and narratives actually spread.

The final dataset analyzed in this study consists of \textbf{81,369} messages, which were subsequently annotated for further investigation.

\subsection{Multi-Class Narrative Assignment}
\label{sec:narratives}
We formulate disinformation detection as a \textbf{multi-class narrative assignment} problem. The narratives dataset is organized into a two-level hierarchy:

\textbf{Narratives (26 total)}: Broad, general themes of disinformation (e.g., ``The West controls Ukraine and uses it for its own purposes'').

\textbf{Sub-narratives (360 total)}: Specific claims and real-world examples that fit within those broader themes (e.g., ``The US is using Ukraine to weaken Europe'').\\
For manual annotation, we sampled 412 messages (\textit{'golden set'}). To capture the corpus's temporal structure and account for activity fluctuations throughout the year, we employed \textbf{stratified random sampling} by week of the year.

Three independent annotators conducted the labeling in two phases. First, to establish inter-annotator agreement (IAA), they labeled \textit{a shared subset} of 103 messages. Second, each annotator independently labeled an additional 103 distinct messages to expand the dataset. Annotators assigned each message to the best-matching narrative from the 26 predefined ones, or marked it as \textit{not containing a narrative}.

During analysis, we observed that many messages contained pro-Russian narratives absent from the original taxonomy. This likely occurred because the messages in the dataset were relatively recent, while the existing set of narratives had been defined earlier and therefore did not fully capture newly emerging disinformation narratives. To address this gap, annotators identified \textbf{20 new sub-narratives}, which were subsequently harmonized and grouped into \textbf{6 new narrative categories}.

Furthermore, because single messages frequently contained multiple contextually related narratives, we grouped these interrelated narratives under broader, overarching meta-narratives.

The final dataset, \textbf{TeleNarratives}, is organized as a three-level taxonomy with \textbf{380 sub-narratives}, \textbf{32 narratives}, and \textbf{9 meta-narratives}. The complete resource, including a \textbf{Neo4j\footnote{\url{https://neo4j.com/}} graph database dump}, is publicly available at \href{https://osf.io/mjcs9/overview?view_only=07148b5ce53b412998ec179587ad8240}{the project repository}.

\subsection{Weak Labeling}
\label{sec:weak_label}
Given the scale of the corpus, full manual annotation was impractical due to limited human resources. Therefore, we adopted a \textbf{weak labeling} approach.

To implement this approach, we explored three different methods. The core idea behind these strategies was to compare each message with the sub-narratives rather than broader narratives. It allows models to perform more precise semantic comparisons between the message content and the narrative descriptions. Our goal was to evaluate the performance of these strategies and select the most effective one based on its agreement with the \textbf{golden set}.

\paragraph{Semantic similarity-based labeling.}
This approach utilizes \textbf{sentence-transformer models} to represent messages and sub-narratives as embedding vectors. Using cosine similarity, each message is assigned to the most similar sub-narrative, provided the score exceeds a predefined threshold. Messages that do not meet this criterion are classified as \textbf{not containing a narrative}.

To generate embeddings, we experimented with three multilingual models: \texttt{paraphrase-multilingual-MiniLM-L12-v2}, \texttt{BAAI/bge-m3}, \texttt{text-embedding-3-large (OpenAI)}\footnote{\url{https://huggingface.co/sentence-transformers/paraphrase-multilingual-MiniLM-L12-v2}} \footnote{\url{https://huggingface.co/BAAI/bge-m3}} \footnote{\url{https://developers.openai.com/api/docs/models/text-embedding-3-large}}.
These models were selected to compare different multilingual embedding approaches, including lightweight open-source models and larger, high-capacity proprietary systems.

\paragraph{Natural Language Inference (NLI)-based labeling.}
Natural Language Inference (NLI) models identify the logical relationship between two text segments as \textit{entailment}, \textit{contradiction}, or \textit{neutral}. In this study, NLI models assess whether a message supports a particular sub-narrative.

Under this framework, the message acts as the \textit{premise}, and the sub-narrative description as the \textit{hypothesis}. The model calculates the probability of entailment for each pair; the message then receives the label of the sub-narrative with the highest score, provided it exceeds a predefined threshold. Messages that do not meet this criterion are classified as \textbf{not containing a narrative}.
We experimented with {\texttt{mDeBERTa-v3-base-xnli-multilingual-nli}}\footnote{\url{https://huggingface.co/MoritzLaurer/mDeBERTa-v3-base-xnli-multilingual-nli-2mil7}} as a multilingual NLI model. We selected this model for its rare Ukrainian language support and strong community validation on Hugging Face.

\paragraph{LLM-based labeling with zero-shot and few-shot prompting.}
In this approach, the model identifies whether a message contains a specific sub-narrative and provides a confidence score. In this sense, the LLM acts as an ‘additional annotator’, producing labels that can be compared with human annotations.

We explored two prompting approaches: \textbf{zero-shot} and \textbf{few-shot}. In the zero-shot setting, the model receives only task instructions. In the few-shot setting, the prompt also includes several labeled examples with brief explanations. These examples show the model how to identify narratives in practice.

To evaluate the performance of this method and identify the optimal price-quality ratio for the task, the following LLMs were selected: \texttt{GPT-4.1}\footnote{\url{https://developers.openai.com/api/docs/models/gpt-4.1}}, \texttt{GPT-4o-mini}\footnote{\url{https://developers.openai.com/api/docs/models/gpt-4o-mini}}, \texttt{Gemini-2.5-flash}\footnote{\url{https://docs.cloud.google.com/vertex-ai/generative-ai/docs/models/gemini/2-5-flash}}, \texttt{Claude-sonnet-4-20250514}\footnote{\url{https://platform.claude.com/docs/en/about-claude/models/overview}}.

\subsection{Share Graph Construction}

We model message diffusion using two complementary layers: (i) an explicit channel-level share graph derived from Telegram forwarding metadata, and (ii) a semantic message-level graph designed to recover repost-like diffusion that is not marked as a native forward.

\paragraph{Forward-based channel share graph.}
Let \(G_{\mathrm{share}}=(V_C,E_F,w_F)\), where \(V_C\) is the set of observed Telegram channels. We add a directed edge \((u \rightarrow v) \in E_F\) when channel \(v\) contains a message forwarded from channel \(u\). Edge weights count observed forwarding events:
\[
w_F(u,v)=\sum_{m \in M_v}\mathbf{1}\!\left[\texttt{fwd\_from\_channel}(m)=u\right],
\]
where \(M_v\) denotes the set of messages posted in channel \(v\). In implementation, we retain a forward edge only when the referenced source message is present in the collected corpus. This restriction avoids links to unobserved sources and ensures that all edges in \(G_{\mathrm{share}}\) are supported by directly observed data.

\paragraph{Semantic message graph for repost-like diffusion.}
Explicit forwarding provides a high-precision, platform-native signal of diffusion, but it does not capture copy-paste reposts or lightly edited message reuse. To approximate such hidden propagation, we construct a message-level semantic graph with edges of type \texttt{SIMILAR\_TO} marked in the dataset.

Before embedding, we normalize message text and exclude duplicates attributable to explicit forwarding chains. We then encode the remaining Ukrainian- and Russian-language messages using \texttt{baai/bge-m3} and retrieve approximate nearest neighbors with an HNSW index \cite{malkov2018efficientrobustapproximatenearest}. 

For each message, we retrieve the top-\(k\) neighbors with \(k=80\), \(M=48\), \(ef_{\mathrm{construction}}=200\), and \(ef_{\mathrm{search}}=200\); these settings preserved recall for subtle multilingual paraphrases while keeping index size and query latency tractable, with only marginal gains beyond them. We retain only high-confidence cross-channel pairs above a calibrated cosine threshold \(\tau=0.985\), excluding same-channel pairs and pairs already linked by explicit forwarding metadata. We selected \(\tau\) using 3{,}215 mapped hidden-forward pairs: recall was 81.0\% at 0.98, 76.39\% at 0.985, and 69.58\% at 0.99, while the number of retained pairs after filtering decreased from 35{,}092 to 24{,}692 and 19{,}095, respectively, providing a practical balance between recall and edge inflation.

We selected \texttt{baai/bge-m3}, \texttt{intfloat/multilingual-e5-large}, and \texttt{gemini-embedding-001} as pilot candidates because all three are strong multilingual embedding models suitable for cross-lingual semantic retrieval. In pilot calibration on manually inspected Ukrainian/Russian message pairs, \texttt{baai/bge-m3} showed the clearest separation among these candidates. In a 16-pair pilot containing exact-like positives and manually identified hard negatives, the hard-negative pairs had a mean cosine similarity of 0.328 under \texttt{baai/bge-m3}, compared with 0.744 for \texttt{intfloat/multilingual-e5-large} and 0.762 for \texttt{gemini-embedding-001}. We therefore used \texttt{baai/bge-m3} for the full embedding run.
\section{Experiments, Evaluation, Results}
\label{sec:experiments}

We evaluate the proposed framework in three stages. First, we measure IAA to verify the reliability of the manually labeled data. Second, we compare the weak labeling strategies from Section~\ref{sec:weak_label} and identify the most effective setup for large-scale narrative assignment. Third, using the selected labels, we analyze the diffusion graphs to study source behavior, cross-group sharing, multi-hop propagation, and narrative-level dissemination patterns. This organization allows us to move from label quality to structural findings and to assess whether combining weak supervision with graph analysis yields an informative view of the spread of disinformation narratives on Ukrainian Telegram.
\subsection{Annotator Agreement}
To evaluate the quality of the manual annotation, we computed \textbf{Cohen’s Kappa}, \textbf{Fleiss’ Kappa}, and \textbf{Krippendorff’s Alpha}. Cohen’s Kappa measures agreement between pairs of annotators while accounting for chance agreement. Fleiss’ Kappa generalizes this measure to multiple annotators and categorical data. Krippendorff’s Alpha provides a flexible measure of inter-annotator reliability that can handle multiple annotators, missing data, and different data types. 

The results are presented in ~\autoref{tab:inter_annotator_agreement}. Binary classification achieved near-perfect agreement 
($\kappa$/$\alpha \approx 0.887$), while meta-narrative labeling reached 
substantial agreement ($\approx 0.719$). Fine-grained narrative annotation showed moderate agreement ($\approx 0.626$), suggesting that this task is at the narrative level relatively subjective and harder to annotate reliably. Notably, all three metrics (Cohen's Kappa, Fleiss' Kappa, and Krippendorff's Alpha) produce almost identical within each category, which strongly confirms the reliability and consistency of these measurements.
\begin{table}[ht]
    \centering
    \begin{adjustbox}{max width=0.5\textwidth}
    \begin{tabular}{lccc}
        \toprule
        \textbf{Metric} & \textbf{Narrative} & \textbf{Meta-narrative} & \textbf{Binary} \\
        \midrule
        Cohen's Kappa (mean) & 0.626 & 0.719 & 0.887 \\
        Fleiss' Kappa        & 0.626 & 0.718 & 0.887 \\
        Krippendorff's Alpha & 0.627 & 0.719 & 0.887 \\
        \bottomrule
    \end{tabular}%
    \end{adjustbox}
    \caption{Inter-annotator agreement results.}
    \label{tab:inter_annotator_agreement}
\end{table}

\vspace{-5pt} 
\subsection{Weak Labeling Results}
The models were executed to evaluate each labeling strategy across the taxonomy. Since each sub-narrative is linked to a specific narrative and meta-narrative, identifying a sub-narrative automatically determines its higher-level categories. This mapping allows predictions to be evaluated at all levels of the hierarchical structure.

Due to label imbalance (i.e., a majority of non-narrative messages and an uneven distribution of specific narratives), we focused on metrics robust to skewed data: \textbf{F1} for binary classification (whether a message contains a narrative), \textbf{Weighted F1} for multi-class classification, \textbf{Matthews Correlation Coefficient (MCC)} for both.

First, we evaluated the models used in the \textbf{semantic similarity}, \textbf{multilingual NLI}, and \textbf{LLM zero-shot prompting} strategies. However, their performance on \textit{the shared subset} of 103 messages was relatively limited (see Table~\ref{tab:weak_labeling_evaluation} for detailed results).

We also explored \textbf{ensemble approaches} that combined predictions from multiple LLMs. In particular, we tested ensembles consisting of \texttt{Claude}, \texttt{GPT-4.1}, and \texttt{Gemini}, as well as \texttt{GPT-4o-mini}, \texttt{GPT-4.1}, and \texttt{Gemini}, assigning a higher weight to Gemini due to its stronger individual performance. However, these ensemble configurations did not lead to a meaningful improvement in overall results (as Gemini consistently outperformed them across all evaluated metrics).

 We found that the main difficulty is that the relationship between a message and a narrative is often \textbf{implicit}. As a result, the message may not be semantically similar to the corresponding narrative description. Instead, the narrative needs to be inferred from the common (sometimes political) context.

Because of this, the \textbf{semantic similarity} and \textbf{multilingual NLI} approaches were not suitable for our task. We therefore modified the LLM-based strategy by using \textbf{few-shot prompting}, adding several annotated examples to the prompt to illustrate the implicit relationship between messages and narratives. This approach produced significantly better results. To ensure that the method performs consistently on a larger sample, we repeated the evaluation on the full set of 412 manually annotated messages.
The results remained strong ($F_1 = 0.82$, $\text{MCC}_{\text{bin}} = 0.71$, $\text{W-}F_1 = 0.76$, $\text{MCC}_{\text{meta}} = 0.57$). 
The final version of the prompt used in the experiments is provided in Appendix ~\ref{sec:appendix_prompts}.

\begin{table}[ht]
\centering
\setlength{\tabcolsep}{3pt} 
\begin{adjustbox}{max width=0.5\textwidth}
\begin{tabularx}{\columnwidth}{X cc cc} 
\toprule
\multirow{2}{*}{\textbf{Classifier}} & \multicolumn{2}{c}{\textbf{Binary}} & \multicolumn{2}{c}{\textbf{Meta}} \\
\cmidrule(lr){2-3} \cmidrule(lr){4-5}
 & F1 & MCC & W-F1 & MCC \\
\midrule
\textit{Semantic Similarity} \\
MiniLM-L12-v2* & 0.51 & 0.14 & 0.41 & 0.11 \\
BGE-M3* & 0.37 & 0.08 & 0.54 & 0.13 \\
OpenAI Text-3* & 0.41 & -0.07 & 0.36 & 0.03 \\
\midrule
\textit{Multilingual NLI} \\
mDeBERTa-v3-base* & 0.54 & 0.15 & 0.32 & 0.13 \\
\midrule
\textit{LLM (Zero-Shot)} \\
GPT-4o-mini* & 0.51 & 0.23 & 0.53 & 0.17 \\
GPT-4.1* & 0.65 & 0.60 & 0.69 & 0.50 \\
Gemini-2.5-Flash* & 0.76 & 0.63 & 0.74 & 0.52 \\
Claude-Sonnet-3.5* & 0.60 & 0.55 & 0.67 & 0.40 \\
\midrule
\textit{LLM (Ensemble)} \\
GPT-4.1 + Claude + Gemini* & 0.69 & 0.61 & 0.72 & 0.50 \\
GPT-4.1 + GPT-4o-mini + Gemini* & 0.71 & 0.61 & 0.70 & 0.44 \\
\midrule
\textit{LLM (Few-Shot)} \\
\textbf{Gemini-2.5-Flash*} & \textbf{0.85} & \textbf{0.78} & \textbf{0.80} & \textbf{0.62} \\
\textbf{Gemini-2.5-Flash**} & \textbf{0.82} & \textbf{0.71} & \textbf{0.76} & \textbf{0.57} \\
\bottomrule
\end{tabularx}
\end{adjustbox}
\caption{LM-based approaches significantly outperform Semantic Similarity and NLI baselines, with \textbf{Gemini-2.5-Flash (Few-Shot)} achieving the highest overall performance. Results marked with (*) denote the \textit{shared subset}, while (**) represents the full \textit{golden set}. Meta denoted Meta-narrative.}
\label{tab:weak_labeling_evaluation}
\end{table}

\subsection{Share Graph Results}

We next analyze the share graph to understand what its structure reveals about how narratives spread across channels. Our analysis focuses on three questions: which channels are likely to occupy upstream positions, how often sharing occurs across channel groups, and how these patterns differ between the explicit-forwarding and semantic-similarity layers. Formal definitions of the metrics and additional robustness details are provided in Appendix~\ref{app:share_graph_details}.

\paragraph{Narrative source detection.}

We first examine whether explicit forwarding structure can identify likely upstream sources of narrative dissemination. For this analysis, a channel is classified as \emph{narrative-active} if it satisfies two conditions: (i) it has at least 50 labeled messages, and (ii) at least 60\% of those labeled messages receive a narrative label. Under this rule, 41 of the 98 observed channels are classified as narrative-active. The remaining 57 channels are treated as non-narrative for this grouping analysis, including 25 channels that meet the minimum label-count requirement but fall below the 60\% narrative threshold, and 32 channels with limited evidence.

To characterize source behavior in the explicit forwarding graph, we use three complementary channel-level metrics: total forwarding volume (\textit{spread\_events}), the number of distinct recipient channels reached (\textit{downstream\_channels}), and a normalized source tendency score (\textit{source\_share}) that distinguishes net sources from channels that mostly relay content from others.

Across the 41 narrative-active channels, we observe 3{,}016 explicit spread events. By forwarding volume, \texttt{rian\_ru} is the dominant source with 612 forwarding events, accounting for \(20.3\%\) of all spread events in this subset. It is followed by \texttt{depzdravzo} (337), \texttt{Pukhov\_M} (303), \texttt{hersonruss} (228), and \texttt{readovkanews} (209). \texttt{rian\_ru} also ranks first in dissemination breadth, reaching 17 distinct downstream channels, and has a source-share score of \(1.00\), indicating a purely source-like position in the observed forwarding network. These results show that the explicit share graph provides a useful signal for identifying likely upstream narrative spreaders, while still reflecting source positions only within the collected network rather than absolute first creators outside it.

\paragraph{Explicit graph hop calibration and sensitivity.}
We selected the hop budget by examining the shortest-path distribution over reachable ordered pairs. The smallest value covering at least \(85\%\) of reachable ordered pairs is \(h=4\), which covers \(92.28\%\) of them.
At \(h=4\), overall cross-group reachability is \(0.0211\), with reachability of \(0.0153\) from narrative-active to non-narrative channels and \(0.0268\) in the reverse direction. Increasing the hop budget to \(h=8\) raises reachability from non-narrative to narrative-active channels to \(0.03\), while reachability from narrative-active to non-narrative channels remains \(0.0153\).

\paragraph{Cross-group sharing analysis.}

We next examine if explicit forwarding crosses channel groups defined by channel-level narrative presence. To quantify this, we compare within-group and cross-group forwarding shares and then examine whether cross-group contact becomes more common when short multi-hop paths are allowed.

Direct cross-group forwarding is rare: among \(34{,}967\) forwarding events, only \(121\) cross group boundaries, corresponding to a share of \(0.0035\). These events are limited in both directions, with \(57\) events from narrative-active to non-narrative channels and \(64\) in the reverse direction. Allowing short multi-hop paths does not substantially change this conclusion. Using a hop budget calibrated on the shortest-path distribution, overall cross-group reachability in the explicit forwarding graph is only \(0.02\), with somewhat higher reachability from non-narrative channels into the narrative-active group than in the reverse direction. Overall, most forwarding remains in-group, and bridge-like cross-group routes are uncommon.

\paragraph{Semantic graph flow and reachability details.}
In the semantic similarity graph, message-level semantic links are oriented by time and aggregated into channel-level semantic flow counts, where \(\mathcal{W}^{\mathrm{sem}}_{u\to v}\) denotes the number of semantic propagation events from channel \(u\) to channel \(v\). Direct cross-group semantic flow is summarized using the same within-group and cross-group share logic as in the explicit forwarding graph. For indirect connectivity, we calibrate the hop budget on the shortest-path distribution over reachable ordered channel pairs. The smallest cutoff covering at least \(85\%\) of reachable ordered pairs is \(h=3\), corresponding to \(85.65\%\) coverage.
At \(h=3\), overall cross-group reachability is \(0.5385\), with reachability of \(0.5049\) from narrative-active to non-narrative channels and \(0.5721\) in the reverse direction. A larger hop budget, \(h=8\), increases these values to \(0.669\) and \(0.6387\), respectively, with \(0.6538\) overall.

\paragraph{Cross-group sharing in the semantic similarity graph.}

To complement the explicit forwarding analysis, we examine cross-group diffusion in the semantic similarity graph, where message-level semantic links are time-oriented and aggregated to channel-level semantic flows. We then compare within-group and cross-group semantic flows and assess whether the two groups are connected via short multi-hop paths.

Direct cross-group semantic flow remains limited but is more common than in the explicit forwarding graph. Among \(52{,}414\) semantic flow events, \(1{,}078\) cross group boundaries, corresponding to a share of \(0.02\). The directional split is also asymmetric: \(256\) events run from narrative-active to non-narrative channels, while \(822\) run in the reverse direction. The main contrast appears in the multi-hop setting. At hop budget \(h=3\), overall cross-group reachability rises to \(0.54\), far above the corresponding value in the explicit forwarding graph. This suggests a repost-like narrative transfer frequently connects the two groups through intermediate channels even when direct cross-group links remain relatively uncommon.

\subsection{Narrative Distribution}
\label{sec:narrative_distr}

\begin{figure}
    \centering
    \includegraphics[width=1\linewidth]{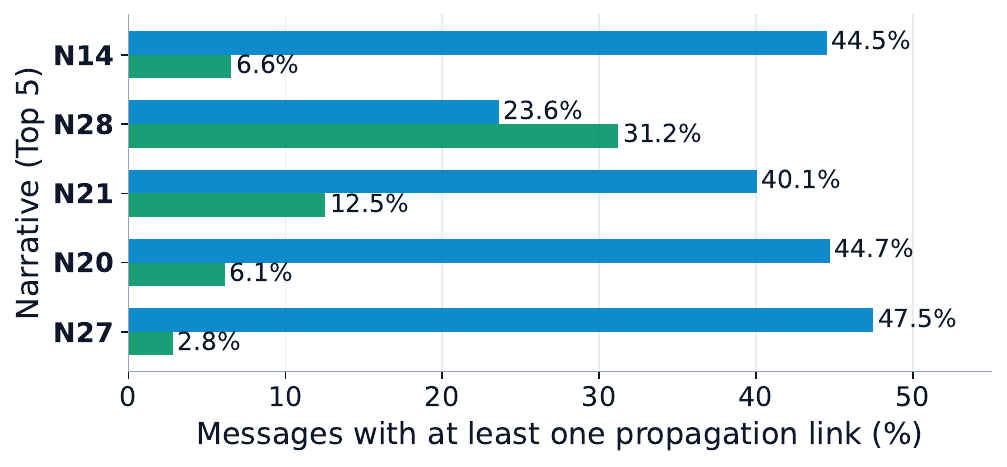}
    \caption{Coverage-based propagation comparison for the five most frequent narratives: N14 (Discrediting or ridiculing representatives of Ukrainian authorities), N28 (Russia improves life in occupied territories), N21 (Ukraine's victory is impossible), N20 (The West controls Ukraine and uses it for its own goals), and N27 (Discrediting the EU and the West). Blue bars show the share of messages with at least one explicit forward (\texttt{FORWARD\_FROM}); green bars show the share of messages with at least one semantic forward (\texttt{SIMILAR\_TO}).}
    \label{fig:narrative-per-share}
\end{figure}

We next examine narrative distribution in the labeled graph scope. Of the \( 81{,}369\) labeled messages, \(29{,}649\) (\(36\%\)) receive a narrative assignment. Figure~\ref{fig:narrative-per-share} compares propagation coverage for these five narratives across the explicit and semantic graph layers. At the macro-family level, the distribution is dominated by \textit{Discrediting Ukraine and its institutions} (\(33\%\)), followed by \textit{Narratives about Russian welfare and ``liberating'' role} (\(17.2\%\)) and \textit{Discrediting the EU and the West} (\(14.2\%\)) (See \autoref{app:narrative_distribution_figures}). This suggests that narrative prominence in the corpus and propagation strength do not align uniformly across diffusion layers (explicit forward and semantic similar forward).
\section{Discussion and Conclusion}
This work introduces a graph-based framework for detecting and analyzing disinformation narratives in Telegram ecosystems by combining weak supervision with propagation graph analysis. Modeling disinformation at the narrative level enables the system to capture semantically related claims that appear in different linguistic forms, addressing the challenge of repeated paraphrasing and reposting common in Telegram channels. The propagation graph further provides a structural perspective on information spread, revealing how narratives circulate across interconnected channels rather than appearing as isolated posts. Graph representation highlights clusters of channels that repeatedly amplify similar narratives, offering insights into how disinformation spreads through platform-specific communication patterns. Beyond improving large-scale detection, this approach provides a framework for studying narrative diffusion and monitoring emerging disinformation campaigns in rapidly evolving online environments such as Telegram, where traditional post-level analysis often fails to capture broader information dynamics.
\section*{Limitations}
\label{sec:limitations}

Our approach has several limitations. First, the framework relies on weak supervision for narrative labeling, which can introduce label noise. While this enables scalable annotation of large Telegram datasets, automatically generated labels may include ambiguous or incorrectly assigned instances, potentially affecting downstream analysis. In the study, we estimated the amount of errors which one can expect from our best-performing solution. However, on different channels and over time the performance might degrade or perform better.

Second, the narrative taxonomy used for labeling may be incomplete and subject to drift over time. As new narratives emerge or existing ones evolve, the predefined taxonomy may fail to capture all relevant claims. In our experiments, we observe indications of narrative drift when additional narratives are introduced, suggesting that narrative boundaries are dynamic and context-dependent.

Third, the dataset is limited to a selected set of Telegram channels, which may introduce sampling bias, especially as we only subsample from the messages that have forwarding connections. The analyzed network may therefore not fully represent the broader Telegram information ecosystem nor the channels where they come from.

Finally, we do not investigate variability of Ukrainian language and code-switch varieties, and how performance of narrative detection decreases on such instances, while this is expected behaviour based on prior research \citet{shynkarov-etal-2025-improving}.
\section*{Ethical Considerations}
\label{sec:ethics}

Our analysis relies on publicly accessible Telegram channels. Although these data are publicly available, users may not anticipate their messages being analyzed in large-scale computational studies. To mitigate potential privacy risks, we focus on aggregate patterns of narrative propagation and big public channels rather than individual user behavior.

Automated systems for detecting disinformation may produce false positives or misclassify legitimate content. Such errors could contribute to unfair labeling of channels or narratives if used without appropriate human oversight. Our framework is intended as a research tool for analyzing information dynamics rather than as a standalone moderation system.

Methods for detecting narrative propagation may also inform adversarial actors about how such systems operate. Increased awareness of detection strategies could encourage actors to adapt their communication patterns to evade analysis.

Using an external narrative inventory (VoxCheck) embeds expert judgments about what constitutes a disinformation narrative. While grounded in professional fact-checking practice, such inventories reflect particular epistemic and institutional perspectives and may not capture all interpretations of contested claims.

\section*{Acknowledgements}
This research was partially supported by ELEKS through a grant dedicated to the memory of Oleksiy Skrypnyk.
The work on this paper is partially performed in the scope of the project “VeraXtract” (16IS24066) funded by the German Federal Ministry for Research, Technology and Aeronautics (BMFTR).

\bibliography{custom}

\clearpage
\appendix
\clearpage
\section{Details for Share-Graph Results}
\label{app:share_graph_details}

\paragraph{Source metrics in the explicit forwarding graph.}
Let \(W_{u \rightarrow v}\) denote the number of explicit forwarding events in which channel \(v\) forwards content originally posted by channel \(u\). We summarize each channel's source behavior using three forwarding-based metrics:
{\small
\[
\begin{aligned}
\textit{spread\_events}(u) &= \sum_{v \neq u} W_{u \rightarrow v},\\
\textit{downstream\_channels}(u) &= \left|\{v \neq u : W_{u \rightarrow v} > 0\}\right|,\\
\textit{source\_share}(u) &=
\frac{\sum_{v \neq u} W_{u \rightarrow v}}
{\sum_{v \neq u} (W_{u \rightarrow v} + W_{v \rightarrow u})}.
\end{aligned}
\]
}
The first metric captures total propagation volume, the second captures dissemination breadth across distinct recipient channels, and the third measures net-source tendency on a bounded \([0,1]\) scale.

\paragraph{Cross-group forwarding metrics.}
To quantify within-group versus cross-group forwarding in the explicit share graph, we compute
{\small
\[
\begin{aligned}
\textit{same\_label\_share} &=
\frac{\sum_{u \neq v} W_{u \rightarrow v}\,\mathbf{1}[y(u)=y(v)]}
{\sum_{u \neq v} W_{u \rightarrow v}},\\
\textit{cross\_label\_share} &= 1-\textit{same\_label\_share}.
\end{aligned}
\]
}
where \(y(u)\) denotes the group label of channel \(u\).

\paragraph{Cross-group reachability definition.}
For ordered channel pairs \((u,v)\), let \(d(u,v)\) be the directed hop distance from \(u\) to \(v\). We define cross-group reachability under hop budget \(h\) as
{\small
\[
\begin{aligned}
R_{\mathrm{cross}}(h) &=
\frac{\left|\{(u,v)\in C : d(u,v)\le h\}\right|}{|C|},\\
C &= \{(u,v): y(u)\neq y(v)\}.
\end{aligned}
\]
}

\section{Cross-Group Shares}
\label{app:cross_group_message_examples}

Table~\ref{tab:cross_group_summary} summarizes the direction of cross-group events, separately for explicit forwarding and semantic similarity.

Additional examples of such messages, along with their detected narratives and sub-narratives, are presented in \autoref{tab:cross_examples_forward_n2n} for cross-group explicit-forwarding from narrative-active to non-narrative channels, \autoref{tab:cross_examples_forward_nn2n} for cross-group explicit-forwarding from non-narrative to narrative-active channels, \autoref{tab:cross_examples_semantic_n2n} for cross-group semantic-similarity from narrative-active to non-narrative channels, and \autoref{tab:cross_examples_semantic_nn2n} for cross-group semantic-similarity from non-narrative to narrative-active channels.

\begin{table}[t]
\centering
\setlength{\tabcolsep}{4pt}
\begin{adjustbox}{max width=0.5\textwidth}
\begin{tabular}{p{0.28\columnwidth}p{0.28\columnwidth}rr}
\toprule
From & To & Events & Share \\
\midrule
\multicolumn{4}{l}{\textbf{Explicit forwarding}} \\
\cmidrule(lr){1-4}
Narrative-active & Non-narrative & 57  & 0.4711 \\
Non-narrative & Narrative-active & 64  & 0.5289 \\
\addlinespace
\bottomrule
\multicolumn{4}{l}{\textbf{Semantic similarity}} \\
\cmidrule(lr){1-4}
Narrative-active & Non-narrative & 256 & 0.2375 \\
Non-narrative & Narrative-active & 822 & 0.7625 \\
\bottomrule
\end{tabular}
\end{adjustbox}
\caption{Summary of cross-group events for explicit forwarding and semantic similarity.}
\label{tab:cross_group_summary}
\end{table}

\section{Chronological List of Events}
\label{app:chronological_events}
A chronological list of events corresponding to markers in Figure \ref{fig:weekly_temporal_analysis} is shown in \autoref{tab:events}.

\begin{table}[ht!]
\centering
\begin{adjustbox}{max width=0.5\textwidth}
\begin{tabularx}{\columnwidth}{@{} l X @{}}
\toprule
\textbf{Date} & \textbf{Event} \\
\midrule
2024-12-19 & EU agreed on a €90B loan programme for Ukraine (2026–2027). \\
2024-12-22 & Slovak PM Fico visits Moscow; protests begin. \\
2025-01-10 & Large-scale protests in Slovakia. \\
2025-01-24 & New protests following "attempted coup" statements. \\
2025-02-28 & Ukraine-US meeting at White House on strategy. \\
2025-03-02 & London Summit on Ukraine. \\
2025-03-21 & German Bundestag approves €3B military aid. \\
2025-06-13 & Iran's massive missile strike on Israel. \\
2025-06-22 & US air strikes on Iranian nuclear facilities. \\
2025-06-23 & Iran attacks US base Al-Udeid in Qatar. \\
2025-07-10 & EU announces €2.3B in reconstruction funding. \\
2025-07-18 & EU adopts 18th package of sanctions. \\
2025-08-25 & EU announces additional €4B in aid. \\
2025-10-23 & EU adopts 19th package of sanctions. \\
2025-10-29 & Updated EU-Ukraine trade agreement in force. \\
2025-12-15 & European leaders propose peace plan. \\
2026-01-06 & Declaration for multinational security force. \\
2026-01-14 & EC proposes €90B in aid for 2026–2027. \\
2026-01-23 & 1st round of negotiations (Abu Dhabi). \\
2026-02-04 & 2nd round of negotiations (Abu Dhabi). \\
2026-02-15 & Munich Security Conference 2026. \\
2026-02-23 & Hungary blocks new EU sanctions/aid. \\
\bottomrule
\end{tabularx}
\end{adjustbox}
\caption{Chronological list of events corresponding to markers in Figure \ref{fig:weekly_temporal_analysis}.}
\label{tab:events}
\end{table}

\section{Narrative Distribution}
\label{app:narrative_distribution_figures}

\begin{figure*}[t]
    \centering
    \includegraphics[width=2\columnwidth]{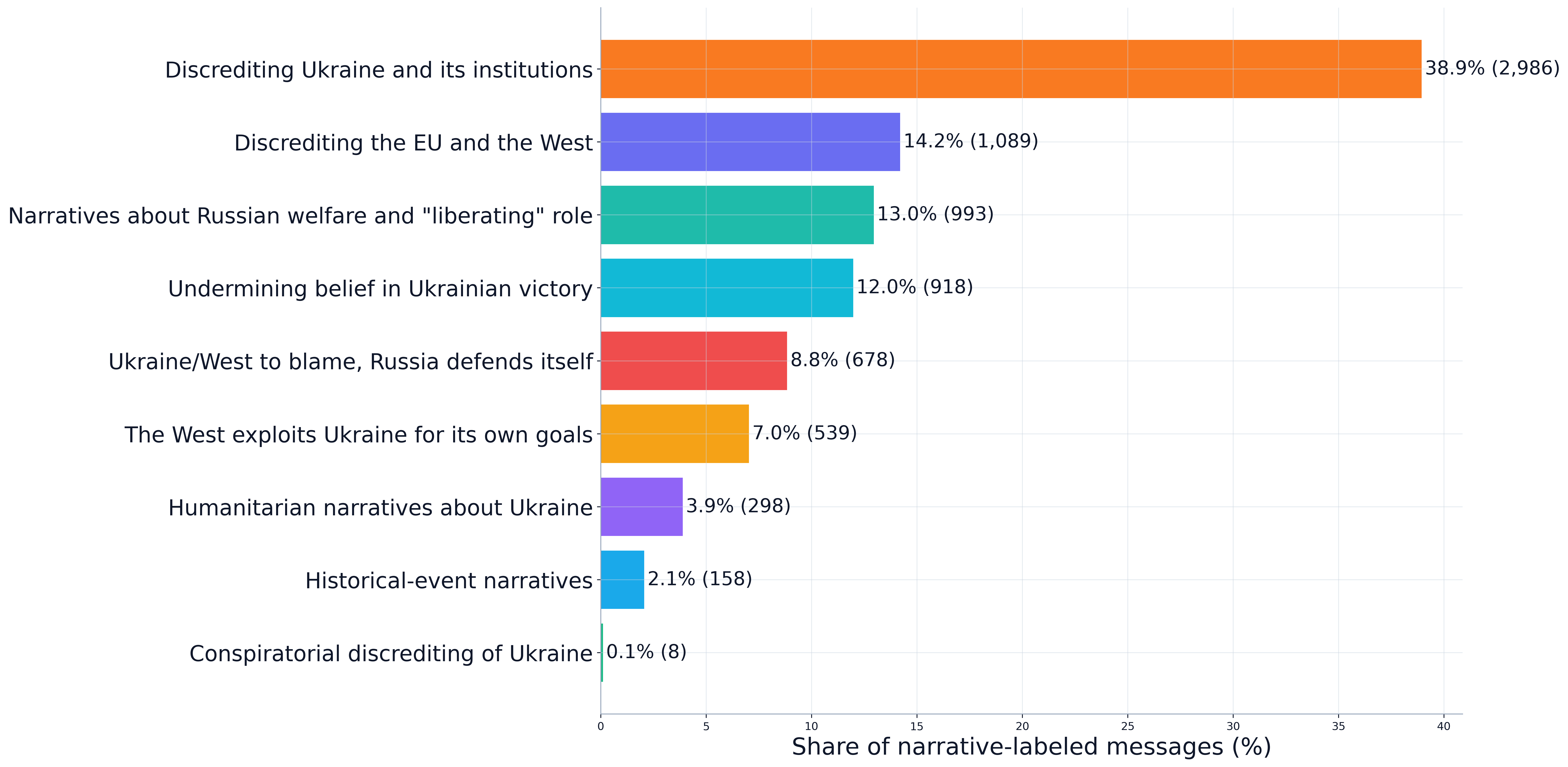}
    \caption{Narrative distribution at the meta-narrative level.}
    \label{fig:app_narrative_distribution_macro}
\end{figure*}

Supplementary visualizations for the narrative distribution analysis discussed in Section~\ref{sec:narrative_distr}. Figure~\ref{fig:app_narrative_distribution_macro} shows narrative distribution at the meta-narrative level.

\begin{table*}[t]
\small
\begin{tabularx}{\textwidth}{X p{0.09\textwidth} p{0.22\textwidth} p{0.22\textwidth}}
\toprule
Message (English translation) & Share count & Narrative & Sub-narrative \\
\midrule
In Kremenchuk, TCC officers chased a young man and deliberately crashed into parked cars. The video shows how the TCC men deliberately ram the car, because they know that the law does not exist for them. A day before that, these same draft officers in uniform opened fire. The head of the Main Directorate of the National Police in Poltava oblast remains silent. & 1 & Discrediting the Ukrainian army & Mobilization in Ukraine violates all standards \\
\midrule
The Wall Street Journal, citing sources in the American administration: the United States has removed the key restriction on Ukraine's use of Western long-range missiles, which will allow Kyiv to strike deep into Russia. This shift coincided with President Trump's attempt in early October to pressure the Kremlin to begin negotiations on ending the war. & 1 & The West controls Ukraine and uses it for its own goals & Ukraine is an instrument of the United States \\
\midrule
Trump has become more detached on the issue of the Ukrainian conflict and no longer lashes out at Russia and Ukraine over the lack of progress in the negotiations, ABC reports, citing American officials. & 1 & Ukraine and the West refuse to start peace negotiations & Ukraine must immediately begin peace negotiations and make compromises with Russia \\
\bottomrule
\end{tabularx}
\caption{Representative cross-group explicit-forwarding examples from narrative-active to non-narrative channels.}
\label{tab:cross_examples_forward_n2n}
\end{table*}

\begin{table*}[t]
\small
\begin{tabularx}{\textwidth}{X p{0.09\textwidth} p{0.22\textwidth} p{0.22\textwidth}}
\toprule
Message (English translation) & Share count & Narrative & Sub-narrative \\
\midrule
It feels as though the wrong people were called occupiers. This is how the TCC tried to abduct a serf in front of his pregnant wife and children. People fought them off, but what moral freaks they are. Ze mobilization is the genocide of the Ukrainian people. & 3 & Discrediting or ridiculing representatives of Ukrainian authorities & Zelensky wants to sacrifice Ukrainians for his own interests \\
\midrule
To call Putin an old fantasist while having 78-year-old Trump next to him is a brilliant move. I love it when this idiot buries himself. & 1 & Discrediting or ridiculing representatives of Ukrainian authorities & Ridiculing representatives of the Ukrainian authorities \\
\midrule
``Ukraine is an artificially created country. When Putin takes Odesa, everything will be fine for us,'' just a survey on the streets of Odesa. More than 3 years of full-scale war. What is in their heads? & 1 & Narrative about historical events & Ukraine developed as an artificial state \\
\bottomrule
\end{tabularx}
\caption{Representative cross-group explicit-forwarding examples from non-narrative to narrative-active channels.}
\label{tab:cross_examples_forward_nn2n}
\end{table*}

\begin{table*}[t]
\small
\begin{tabularx}{\textwidth}{X p{0.09\textwidth} p{0.22\textwidth} p{0.22\textwidth}}
\toprule
Message (English translation) & Share count & Narrative & Sub-narrative \\
\midrule
War has been declared today on the Russian world --- Vladimir Putin. & 2 & Actions of Ukraine and the West forced Russia to start the war & Russia was forced to enter the war to ensure its survival \\
\midrule
Putin supported Trump's idea of a mutual refusal by Russia and Ukraine for 30 days to strike energy infrastructure and gave such an order to the military --- Kremlin. & 2 & Ukraine and the West refuse to start peace negotiations & Unlike Ukraine, Russia demonstrates readiness for negotiations \\
\midrule
Poland does not want simply to be in solidarity with Ukraine; it wants to profit from it. ``We will no longer help in a naive way. It will not be that Poland is solidaristic and others profit from the reconstruction of Ukraine,'' said Polish Prime Minister Tusk. & 2 & The West controls Ukraine and uses it for its own goals & Western elites are the main beneficiaries of the war \\
\bottomrule
\end{tabularx}
\caption{Representative cross-group semantic-similarity examples from narrative-active to non-narrative channels.}
\label{tab:cross_examples_semantic_n2n}
\end{table*}

\begin{table*}[t]
\small
\begin{tabularx}{\textwidth}{X p{0.09\textwidth} p{0.22\textwidth} p{0.22\textwidth}}
\toprule
Message (English translation) & Share count & Narrative & Sub-narrative \\
\midrule
In Pokrovsk, almost 15 million were allocated to support the media, 73 million to the water utility, 34 million to beautification, and almost 30 million to the functioning of heat energy, --- Telegraf, citing the community budget for 2025. Fifty million is provided for officials' salaries, 98 million for the elimination of emergency situations, and 147 million for education. According to DeepState, the Russian Armed Forces are already 2--3 km from the city. As of the end of the year, 70\% of residential buildings, 80\% of social facilities, and 95\% of industrial facilities had been damaged or destroyed. There is no electricity, gas, heating, or water supply in the city.& 3 & Discrediting or ridiculing representatives of Ukrainian authorities & Corruption in Ukraine's political and military leadership \\
\midrule
Slovakia threatens Ukraine with stopping electricity supplies because of the halt in gas supplies. ``If necessary, we will stop electricity supplies, which Ukraine critically needs during power outages,'' Fico said. & 2 & Ukraine is left alone and will lose & Partners abandoned Ukraine \\
\midrule
The Biden administration informed Trump's team that Ukraine will have to ``resolve the issue of lowering the mobilization age,'' --- Sullivan. In his opinion, the shortage of people remains an acute problem while the United States is providing ``a huge amount of ammunition and military equipment.'' & 2 & The West controls Ukraine and uses it for its own goals & Ukraine is an instrument of the United States \\
\bottomrule
\end{tabularx}
\caption{Representative cross-group semantic-similarity examples from non-narrative to narrative-active channels.}
\label{tab:cross_examples_semantic_nn2n}
\end{table*}

\section{Final LLM Prompt}
\label{sec:appendix_prompts}
This appendix provides both the original Ukrainian version (Table ~\ref{tab:final_prompt_ukr}) and the English translation (Table ~\ref{tab:final_prompt_eng}) of the final prompt used in the LLM-based weak labeling approach. The prompt was executed using \texttt{Gemini-2.5-flash}.
\begin{table*}[t]
\begin{tabularx}{\textwidth}{p{0.15\textwidth} X}
\toprule
 & \textbf{Ukrainian version} \\
\midrule
System \newline instructions & 
\foreignlanguage{ukrainian}{Ти уважний український класифікатор. Поверни ЛИШЕ валідний JSON. Без зайвого тексту. НЕ вигадуй. НЕ використовуй markdown. НЕ перекладай і не перефразовуй текст наративів. У відповіді повертай ТІЛЬКИ narrative\_id та sub\_narrative\_id (без тексту наративів).} \\
\midrule
Task \newline instructions & 
\foreignlanguage{ukrainian}{
Завдання: Визнач, чи повідомлення ПРЯМО або НЕПРЯМО ('натякаючи') просуває будь-який наратив зі списку.
\newline
Правила:
\begin{itemize}[noitemsep, topsep=2pt]
    \item Спочатку виріши, чи є хоча б один відповідний наратив.
    \item Відповідай «так» лише якщо твердження у повідомленні прямо підтримує наратив.
    \item Якщо це загальні новини або нейтральний опис подій — НЕ став жодних наративів.
    \item Може бути таке, що повідомлення може містити кілька наративів. Обери той, що найбільше підходить, або найбільш чіткий.
    \item Читай уважно. Дивись на повідомлення з української перспективи.
    \item Використовуй ТІЛЬКИ narrative\_id та sub\_narrative\_id зі списку.
    \item Якщо жоден наратив не підходить — поверни null для narrative\_id і sub\_narrative\_id та confidence 0.
    \item Додай confidence від 0 до 1.
    \item Вивід має бути валідним JSON.
    \item Наприклад:
    \begin{itemize}
        \item "кличко заявив що україна стикається з браком солдатів і пропонує знизити вік мобілізації деталі за посиланням" \newline – це повідомлення нав'язує думку, що Кличко хоче знизити мобілізаційний вік. Це страшить громадян і у них виникає погана думка про політика Кличка, тому тут є наратив "Дискредитація чи висміювання представників української влади"
        \item "мужики красавчики понимая что ударные дроны вылетают з контейнера на прицепе фуры начали забрасывать его камнями рискуя своей жизнью конечно они понимали об опасности было бы хорошо если бы их нашли и наградили просто красавчики neoficialniybezsonov" \newline – повідомлення стимулює 'захоплюватись' сміливістю росіян і хоче переконати, що перемога росіян однозначна, тому тут присутній наратив "Перемога України неможлива"
    \end{itemize}
\end{itemize}
} \\
\bottomrule
\end{tabularx}
\caption{Original version of final prompt.}
\label{tab:final_prompt_ukr}
\end{table*}
\begin{table*}[t]
\centering
\begin{tabularx}{\linewidth}{p{0.15\textwidth} X}
\toprule
& \textbf{English version} \\
\midrule
System \newline instructions & You are a diligent Ukrainian classifier. Return ONLY valid JSON. No extra text. Do NOT make anything up. Do NOT use Markdown. Do NOT translate or paraphrase the narrative text. In your response, return ONLY the narrative\_id and sub\_narrative\_id (without the narrative text). \\
\midrule
Task \newline instructions &
Task: Determine whether the message DIRECTLY or INDIRECTLY (‘by implication’) promotes any of the narratives listed below. \newline
Rules:
\begin{itemize}[noitemsep]
    \item First, decide whether there is at least one relevant narrative.
    \item Answer ‘yes’ only if the statement in the post directly supports the narrative.
    \item If it is general news or a neutral description of events — DO NOT assign any narratives.
    \item It may be the case that a post contains several narratives. Choose the one that is most appropriate or the clearest.
    \item Read carefully. View the post from a Ukrainian perspective.
    \item Use ONLY narrative\_id and sub\_narrative\_id from the list.
    \item If no narrative is suitable — return null for narrative\_id and sub\_narrative\_id and confidence 0.
    \item Set confidence to a value between 0 and 1.
    \item The output must be valid JSON.
    \item For example:
    \begin{itemize}
        \item "Klitschko has stated that Ukraine is facing a shortage of soldiers and is proposing to lower the conscription age – details via the link" \newline – this message seeks to suggest that Klitschko wants to lower the conscription age. This frightens citizens and gives them a negative impression of the politician Klitschko; therefore, there is a narrative "Discrediting or ridiculing representatives of the Ukrainian authorities"
        \item "A bunch of cool guys, realising that attack drones were taking off from a container on the back of a lorry, started pelting it with stones, risking their lives. Of course, they knew the danger involved. It would have been great if they’d been found and rewarded – just a bunch of cool guys. neoficialniybezsonov" – The post encourages people to ‘marvel’ at the Russians’ bravery and aims to convince them that a Russian victory is inevitable, so the narrative here is "A Ukrainian victory is impossible"
    \end{itemize}

\end{itemize} \\
\bottomrule
\end{tabularx}

\caption{English version of the prompt used for weak labeling (translated from Ukrainian via DeepL). To ensure reproducibility, please refer to the original Ukrainian prompt in the Table \ref{tab:final_prompt_ukr}.}

\label{tab:final_prompt_eng}
\end{table*}

\end{document}